\documentclass[letterpaper]{article} 
\usepackage{aaai23}  
\usepackage{times}  
\usepackage{helvet}  
\usepackage{courier}  
\usepackage[hyphens]{url}  
\usepackage{graphicx} 
\usepackage{natbib}  
\usepackage{caption} 
\usepackage{algorithm}
\usepackage{algorithmic}

\usepackage{ifthen}
\usepackage{color}
\usepackage{multirow}
\usepackage{mathtools}
\usepackage{bigstrut}
\DeclarePairedDelimiter{\norm}{\lVert}{\rVert}
\DeclarePairedDelimiter\abs{\lvert}{\rvert}%

\definecolor{frenchblue}{rgb}{0.0, 0.45, 0.73}
\definecolor{gray}{rgb}{0.5,0.5,0.5} 
\definecolor{green}{rgb}{0, 0.4, 0}  
\definecolor{orange}{rgb}{1, 0.5, 0}    
\definecolor{mahogany}{rgb}{0.75, 0.25, 0.0}
\definecolor{purple}{rgb}{0.6, 0, 0.6}
\definecolor{darkgreen}{rgb}{0, 0.4, 0.4} 
\definecolor{teal}{rgb}{0.0, 0.5, 0.5}
\definecolor{aaaa}{rgb}{0.55, 0.1, 0.7}
\definecolor{red}{rgb}{1.0, 0, 0}

\newboolean{revising}
\setboolean{revising}{false}
\ifthenelse{\boolean{revising}}
{
    \newcommand{\fuen}[1]{\textcolor{blue}{[FuEn]: #1}}
    
} {
    \newcommand{\fuen}[1]{#1}
    
}

\usepackage{newfloat}
\usepackage{listings}
\DeclareCaptionStyle{ruled}{labelfont=normalfont,labelsep=colon,strut=off} 
\floatstyle{ruled}
\newfloat{listing}{tb}{lst}{}
\floatname{listing}{Listing}
\title{MixFairFace: Towards Ultimate Fairness via MixFair Adapter in Face Recognition}
\author{
    Fu-En Wang\textsuperscript{\rm 1,\rm 2},
    Chien-Yi Wang\textsuperscript{\rm 1},
    Min Sun\textsuperscript{\rm 2},
    Shang-Hong Lai\textsuperscript{\rm 1,\rm 2}
}
\affiliations{
    \textsuperscript{\rm 1}Microsoft AI R\&D Center, Taiwan\\
    \textsuperscript{\rm 2}National Tsing Hua
University, Taiwan\\
    fulton84717@gapp.nthu.edu.tw, chienyiwang0922@gmail.com, sunmin@ee.nthu.edu.tw, lai@cs.nthu.edu.tw
}

\begin{document}

\maketitle
\newcommand{\MA}[0]{\text{MixFair Adapter}}
\begin{abstract}
Although significant progress has been made in face recognition, demographic bias still exists in face recognition systems. For instance, it usually happens that the face recognition performance for a certain demographic group is lower than the others. In this paper, we propose MixFairFace framework to improve the fairness in face recognition models. First of all, we argue that the commonly used attribute-based fairness metric is not appropriate for face recognition. A face recognition system can only be considered fair while every person has a close performance. Hence, we propose a new evaluation protocol to fairly evaluate the fairness performance of different approaches. Different from previous approaches that require sensitive attribute labels such as race and gender for reducing the demographic bias, we aim at addressing the identity bias in face representation, i.e., the performance inconsistency between different identities, without the need for sensitive attribute labels. To this end, we propose \MA~to determine and reduce the identity bias of training samples. Our extensive experiments demonstrate that our MixFairFace approach achieves state-of-the-art fairness performance on all benchmark datasets.
\end{abstract}
\section{Introduction}
\label{sec:intro}
\begin{figure}[t]
    \centering
    \includegraphics[width=\columnwidth]{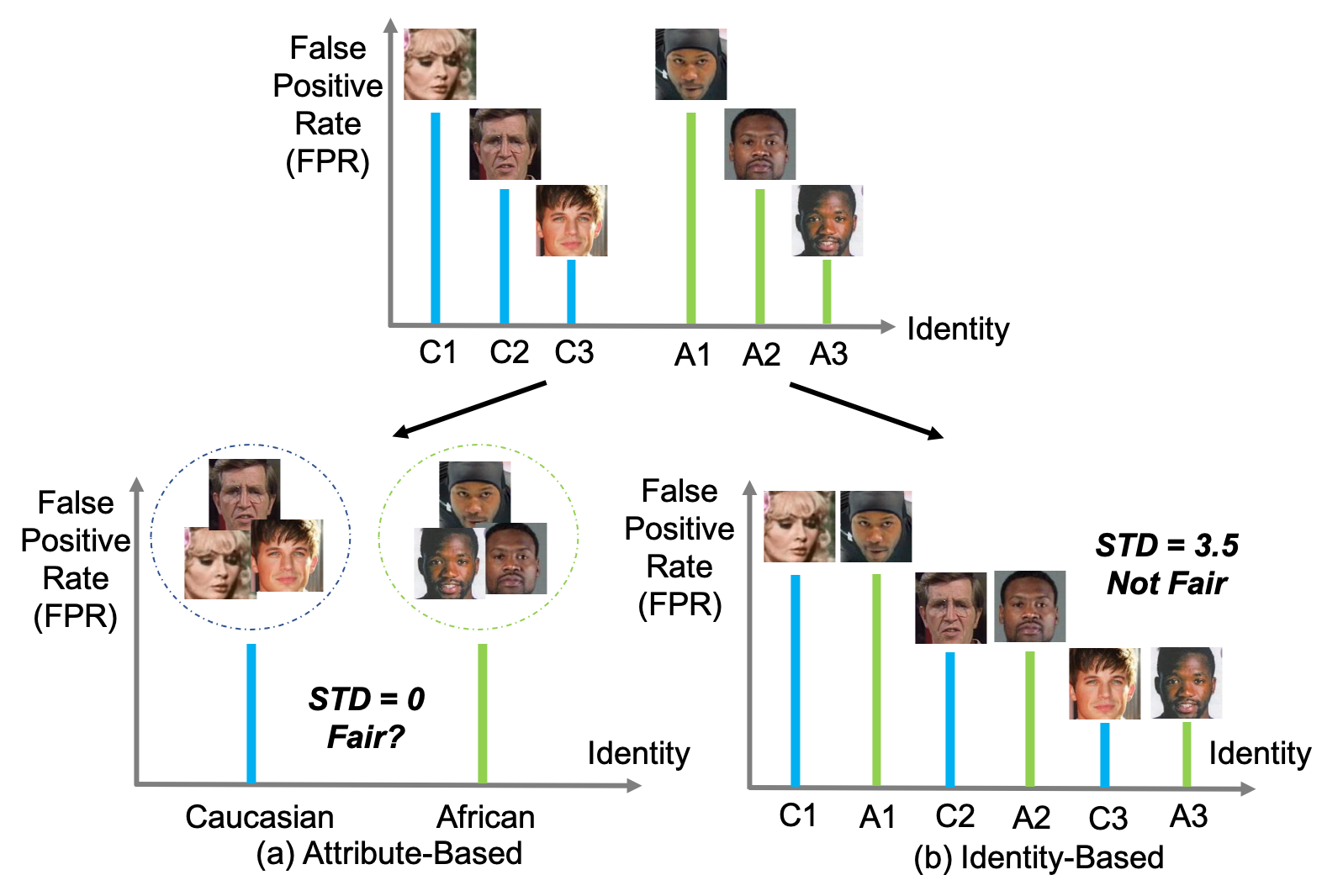}
    \caption{For face recognition, the attribute-based (e.g., race) fairness metric is not enough to represent the bias level in the face representation. The above example demonstrates that although the false positive rate (FPR) is the same for African and Caucasian, there can be high skewness between each identity regardless of their race. In this work, we thus aim to reduce the variance of identity-based FPR to pursue ultimate fairness in face recognition.}
    \label{fig:teaser}
\end{figure}

With the rise of responsible AI in recent years, ethnicity issues such as demographic bias or fairness start to be discovered in face recognition systems~\cite{cifp, pass}. It is a common phenomenon that the performance of deep neural networks is largely inconsistent across different demographic groups such as race and gender, and several approaches for solving demographic bias in face recognition have been proposed in recent years. For instance, \cite{rl-rbn} adopted reinforcement learning to learn a policy that automatically adjusts the margin of loss functions to balance the performance inconsistency between each demographic group. 
\cite{debface,pass} adopted an adversarial training to suppress sensitive attribute features. 
However, as addressed in \cite{cifp}, most existing approaches require sensitive attribute labels (e.g., race and gender) for training the networks, which restricts the scalability for being extended to large-scale datasets since sensitive attributes need to be accurately annotated by humans. In addition to the requirement of sensitive attributes, our experimental results show that the evaluation process adopted by existing works cannot precisely demonstrate the real fairness distribution of each demographic group. In other words, we found the previous evaluation protocol highly underestimates the bias between different racial groups on benchmark datasets.

To precisely evaluate the bias between different racial groups, we propose a new evaluation protocol which consists of both attribute-based and identity-based metrics. For attribute-based metrics, we focus on evaluating the performance inconsistency between different races. As illustrated in Figure~\ref{fig:teaser}, we found that the attribute-based metrics usually raise a fairness ``skewness'' problem even if the performance inconsistency between races is small. Hence, our identity-based metrics aim to measure the ``identity bias'', which is defined as the performance variance between ``each identity''. In this way, we can prevent the skewness issue and consider everyone fairly to achieve ultimate fairness without dividing people by their races.

In addition to the proposed evaluation protocol, we present our ``MixFairFace'' framework to alleviate the identity bias of networks without accessing sensitive attributes. Unlike previous approaches that focus on the bias of demographic groups, we aim at solving the identity bias, the performance inconsistency between different identities, by reducing the feature discriminability differences. To this end, we propose ``\MA'' which estimates the identity bias difference between two identities. The estimated bias difference is then minimized by our loss functions. In this way, we can achieve ultimately fair results since the identity biases in the training dataset can be well balanced.

We demonstrate the effectiveness of MixFairFace on several benchmark datasets. We adopt Balancedface and Globalface~\cite{rl-rbn} as our training datasets, and we use Racial Faces in the Wild~\cite{RFW} for evaluation. In addition, we extend our framework to larger datasets, including MS1M~\cite{guo2016ms} and IJB-C~\cite{IJB-C}. From our extensive experiments, our proposed MixFairFace framework achieves state-of-the-art fairness performance. Our source code and supplementary material are released in our project website\footnote{https://github.com/fuenwang/MixFairFace}.

To summarize, our contributions are listed as follows:
\begin{enumerate}
    \item We propose a new evaluation protocol to more precisely evaluate the fairness in face recognition compared with the previous evaluation process.

    \item We propose \MA~to estimate and reduce the identity bias between different identities during training.

    \item We propose MixFairFace framework to alleviate the identity bias of networks and achieve state-of-the-art fairness in face recognition on several benchmark datasets.
\end{enumerate}

\section{Related Works}
\subsection{General Face Recognition}
In recent years, advanced face recognition approaches have been proposed~\cite{schroff2015facenet, he2016deep, vaswani2017attention, wang2018cosface, deng2019arcface}, and large-scale datasets are collected for training purpose~\cite{cao2018vggface2, guo2016ms, zhu2021webface260m, rl-rbn}.
Among these factors, the loss design is the most pivotal. By adopting strict margin constraints on the positive logits, the model can learn discriminative representation which is able to generalize to unseen identities. However, these margin-based softmax loss functions do not consider the fitting difficulties between different identities and inevitably produce biased representation.

\subsection{Bias Mitigation in Face Recognition}
With the rising awareness of responsible AI, many empirical studies~\cite{nistfrvt, bansal2017s, drozdowski2020demographic} have shown that demographic bias exists in many publicly available face recognition systems. They demonstrated that non-white or darker skintone faces suffer from higher false positive rates (FPR) than other groups. Analysis in ~\cite{gwilliam2021rethinking} also indicates that biases in face recognition models are not mainly caused by imbalanced datasets. Therefore, tailored methods are required to address fairness issues in face recognition.

In order to reduce face recognition biases between racial groups, ~\cite{RFW, zhu2022local} proposed to leverage domain adaptation to produce an optimal model for each race. However, these approaches require unlabeled data from target domains for adaptation and it is not practical in many scenarios. Another line of approaches~\cite{pass, faircal} attempts to remove bias from a pre-trained face recognition model by building a fairer decision system: PASS~\cite{pass} presented a novel discriminator training strategy that discourages the face descriptor from encoding gender and skintone information, but it has the disadvantage of heavily relying on sensitive attribute labels from the face recognition dataset. FairCal~\cite{faircal} proposed to post-calibrate the verification score between each pair of images in the testing dataset. However, this method has the severe limitation of requiring the global statistics of the target face features, which is usually unknown in real scenarios.

Most of the state-of-the-art methods~\cite{rl-rbn, gac, debface, cifp} and our MixFairFace lie in the last category: learning a fair face representation in an end-to-end fashion. DebFace~\cite{debface} proposed an adversarial learning framework to learn disentangled representation for both face recognition and demographics estimation. In the following work~\cite{gac}, they leveraged Group Adaptive Classifier (GAC) to learn different network parameters based on the demographic information. RL-RBN~\cite{rl-rbn} adopted reinforcement learning to find optimal angular margins for African, Indian, and Asian. It also requires race labels in the training data for searching parameters. However, as shown in ~\cite{gwilliam2021rethinking} and our experiments, the boundary between different races is vague and the estimation from a pre-trained race classifier is not accurate. Unlike the above works, our proposed MixFairFace does not require a pre-trained demographic classifier or demographic labels in the training data for model training, which can be generalized to diverse scenarios. Our MixFairFace is inspired from CIFP~\cite{cifp}, which balances the false positive rate (FPR) between each training sample without the need for demographic labels.

\section{Evaluation Protocol}
\label{sec:eval}

As described in Section~\ref{sec:intro}, we found the evaluation process adopted by previous works highly underestimates the bias between demographic groups. Most previous works evaluate fairness performance on Racial Faces in the Wild (RFW)~\cite{RFW} dataset. The dataset consists of four races (African, Asian, Caucasian, and Indian) and each of them contains about 3,000 identities. To evaluate the performance of different races, RFW provides 6,000 verification pairs for each race, and thus there are 24,000 verification pairs in total. With the provided pairs, previous works~\cite{RFW,rl-rbn,pass,gac,cifp} calculate the corresponding verification accuracy of four races and use the standard deviation of four accuracy values as the fairness performance (e.g. Table~\ref{tab:cosface-baseline}). Generally, we found this protocol has the following four issues:
\begin{enumerate}
    \item The 6,000 verification pairs are selected according to their cosine similarity from a baseline model to avoid performance saturation. However, such a selection directly introduces human biases to the verification pairs, and we found this selection leads to the underestimation of demographic bias in RFW.
    
    \item The provided verification pairs only cover a small ratio of the entire dataset, which is not representative enough to the distribution of four races.
    
    \item The verification pairs only consider intra-race comparison, i.e., the comparison of identities in the same race, and inter-race comparison is also necessary for face recognition systems.
    
    \item The four races are evaluated independently, and thus the verification thresholds of the four races are totally different. However, the verification threshold should be set globally in face recognition systems for generalized usage.
\end{enumerate}

\subsection{Intra/Inter-Identity Analysis}
\begin{figure}
    \centering
    \includegraphics[width=\columnwidth]{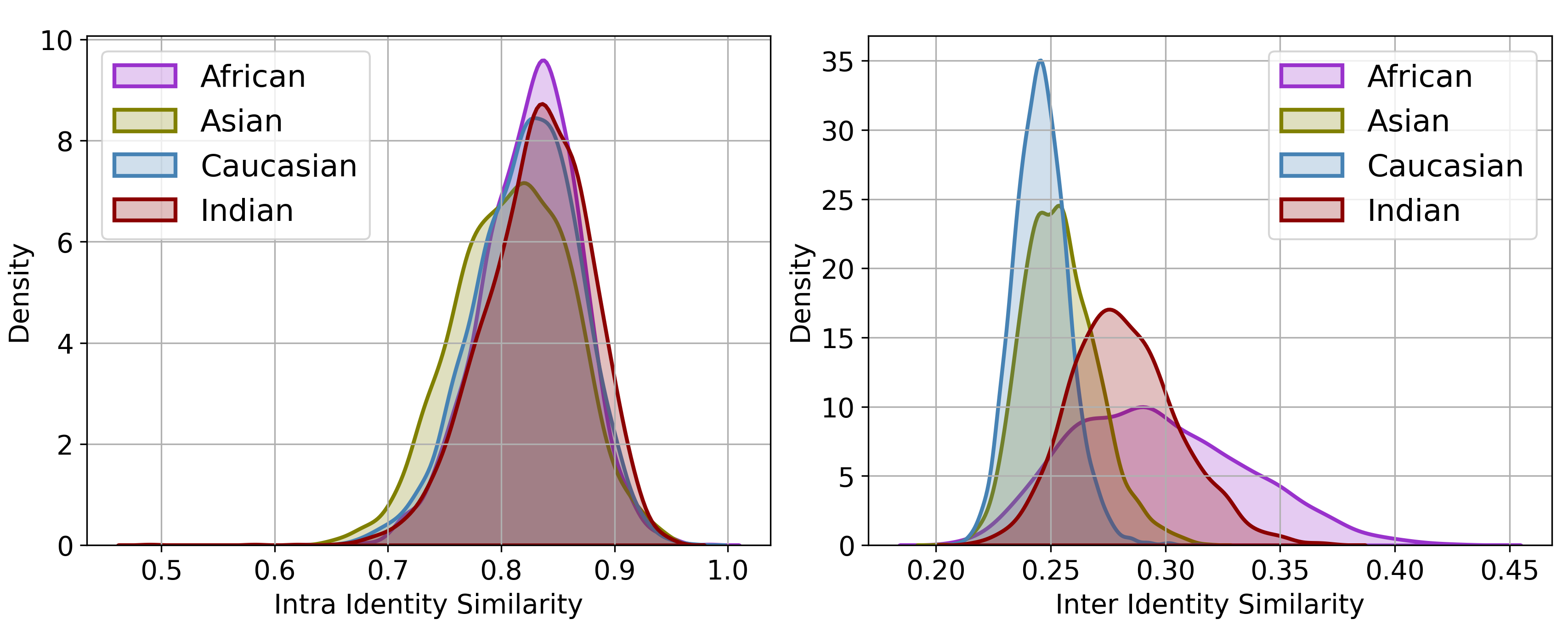}
    \caption{The intra/inter-identity similarity distribution of CosFace~\cite{wang2018cosface}. Comparing with intra-identity similarity (left), we found that there is an obvious difference in inter-identity similarity (right) between races, and the inter-identity similarity is directly related to the overall false positive rate of each race.}
    \label{fig:cosface-intra-inter}
\end{figure}

\begin{table}[t]
  \centering
  \caption{The performance of Cosface model~\cite{wang2018cosface} trained on Balancedface~\cite{rl-rbn}. The accuracy of four races is calculated by the evaluation protocol provided by RFW~\cite{RFW}. We found this evaluation protocol highly underestimates the biases between each race compared with the inter-identity similarity distribution in  Figure~\ref{fig:cosface-intra-inter} (right).}
  \begin{tabular}{|c|c|c|c|c|c|}
    \hline
    \textbf{African} & \textbf{Asian} & \textbf{Caucasian} & \textbf{Indian} & \textbf{Avg}   & \textbf{Std} \bigstrut\\
    \hline
    \hline
    94.1     & 94.2     & 96.3     & 94.8     & 94.9     & 1.03 \bigstrut\\
    \hline
  \end{tabular}%
  \label{tab:cosface-baseline}%
\end{table}%

To address the first and second issues, we first adopt a CosFace~\cite{wang2018cosface} model trained on Balancedface~\cite{rl-rbn} as a baseline model and we extract the feature vectors of all facial images in RFW, which produces about 40,000 feature vectors. Then, we calculate the mean feature vector of each identity in RFW. We use the mean vectors to calculate the distribution of intra-identity and inter-identity cosine similarity:
\begin{equation}
    \begin{aligned}
        S^{intra}_i &= \frac{1}{n_i} \sum_{j=1}^{n_i} \cos(f_j^i, m_i)~, \\
        S^{inter}_i &= \frac{1}{K} \sum_{j\in N_i} \cos(m_j, m_i)~,
    \end{aligned}
    \label{eq:similarity}
\end{equation}
where $i$ denotes the identity, $n_i$ is the total number of images belonging to identity $i$, $m_i$ denotes the mean feature vector for identity $i$, $f_j^i$ is the $j$-th feature vector for identity $i$, and $N_i$ is the set of the closest $K$ mean feature vectors w.r.t. $m_i$ in the dataset, which is used for calculating the inter-identity similarity $S^{inter}_i$. In this paper, we set $K$ to 50.

With the calculated intra/inter-identity similarity for each identity, we can analyze the corresponding distribution as shown in Figure~\ref{fig:cosface-intra-inter}. Interestingly, we found the intra-identity similarity difference between races is small, while there are very large variances in inter-identity similarities between races and these variances reflect the corresponding demographic biases. In general, the inter-identity similarities of African and Indian are the highest, and this is inconsistent with the results of the original evaluation protocol (Table~\ref{tab:cosface-baseline}). Hence, we believe that the original protocol highly underestimates the bias between demographic groups. 

Instead of using the provided verification pairs, we use all facial images in datasets for evaluation. We first calculate the similarity of all pair-wise combinations of facial images in datasets. Considering that a low false positive rate (FPR) is necessary for face recognition systems, we first find the verification threshold under a certain overall FPR of all pair-wise combinations. In this paper, we choose 1e-5 as the overall FPR. After finding the verification threshold, we find the corresponding true and false positive rate (TPR/FPR) of each identity by accumulating their pair-wise verification predictions. We then propose attribute-based and identity-based fairness metrics to more precisely evaluate the fairness performance in face recognition.

\begin{enumerate}
    \item \textbf{Attribute-based fairness evaluation.~} We calculate the average and standard deviation of TPRs and FPRs which belong to a certain sensitive attribute. These metrics can show the performance variance between different demographic groups. For simplicity, we use ``aTPR'' and ``aFPR'' to denote the metrics.
    
    \item \textbf{Identity-based fairness evaluation.~} Our identity-based metric is to measure the identity bias, which is defined as the performance variance across different identities. Hence, we use the standard deviation across the average FPR of each identity as the metric since FPRs of different identities vary much more significantly than other metrics~\cite{cifp}. We use this metric to reflect the fairness of entire datasets and avoid the skewness of attribute-based evaluation as described in Figure~\ref{fig:teaser}. For simplicity, we use ``iFPR-std'' to denote this metric.
\end{enumerate}
\fuen{For more details and definitions of the two metrics, please refer to our supplementary material.}

\begin{figure*}[t]
    \centering
    \includegraphics[width=\textwidth]{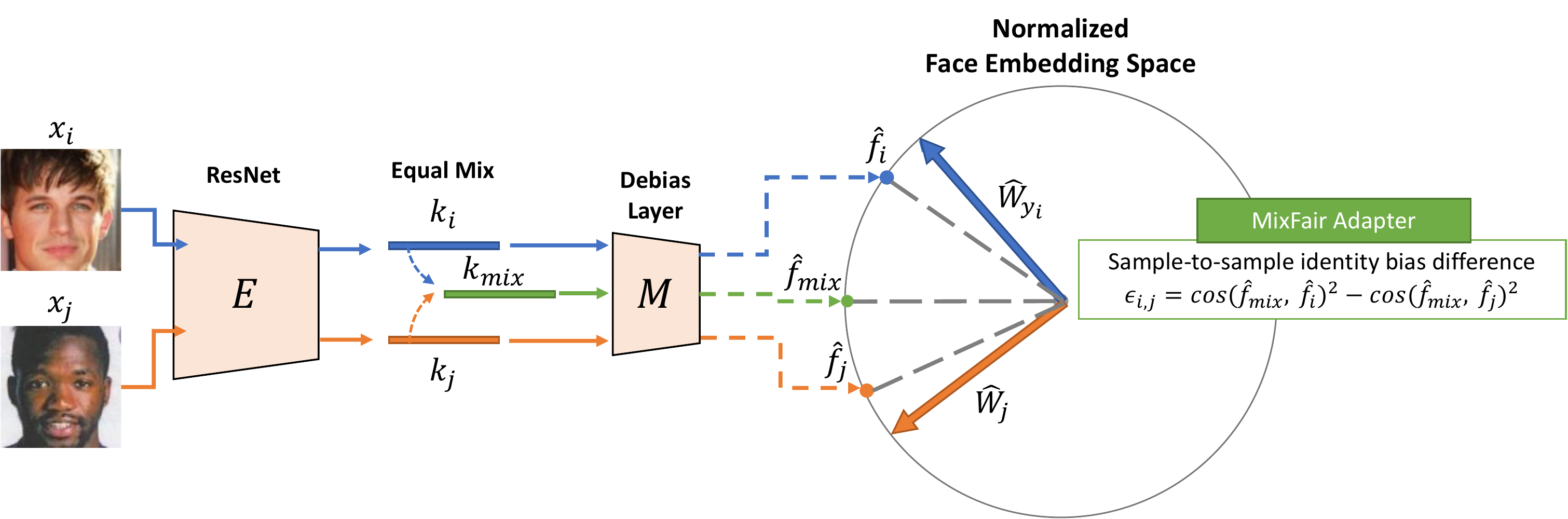}
    \caption{The overall framework of MixFairFace. We first adopt a ResNet encoder $E$ to extract the intermediate feature maps of two identities ($k_i$ and $k_j$). Then, we equally mix the two feature maps and obtain the mixed feature map $k_{mix}$. The original and mixed feature maps ($k_i$, $k_j$, and $k_{mix}$) are further passed into our debias layer $M$ along with a normalization operation to infer the final normalized feature vectors ($\hat{f}_i$, $\hat{f}_j$, and $\hat{f}_{mix}$). We then use our proposed \MA~to estimate the identity bias difference $\epsilon_{i,j}$ between the two training samples ($x_i$ and $x_j$) by calculating the distance between $\hat{f}_i$, $\hat{f}_j$, and $\hat{f}_{mix}$. The estimated bias difference is then minimized by our final loss function (Equation~\eqref{eq:our-loss}) for learning fair representations.}
    \label{fig:approach}
\end{figure*}
\section{Approach}
In this section, we introduce our MixFairFace framework with the proposed components. We first introduce the prototype-based loss function in Section \ref{sec:softmax-loss}. To reduce the identity bias and improve fairness, we use our proposed \MA~(Section~\ref{sec:mixfair}) to estimate the identity bias during training. Finally, we detail our MixFairFace framework in Section~\ref{sec:mixfairface}.

\subsection{Prototype-based Loss}
\label{sec:softmax-loss}
The original softmax loss function commonly adopted in prototype-based face recognition approaches is:
\begin{equation}
    \begin{aligned}
        L = -\log \frac{e^{W_{y_i} \cdot f_i + b_{y_i}}}{\sum_{j=1}^N e^{W_j \cdot f_i + b_j}}~,
    \end{aligned}
    \label{eq:softmax-loss}
\end{equation}
where $N$ denotes the number of identities in the training dataset, $W$ and $b$ are the weight and bias terms of the classification layer, $f_i$ is the feature map of training sample $i$, and $y_i$ is the corresponding ground truth identity. By fixing the bias terms to zero and applying normalization to $W$ and $f_i$, the exponential parts in Equation~\eqref{eq:softmax-loss} can be reformulated as:
\begin{equation}
    \begin{aligned}
        \hat{f}_i &= \frac{f_i}{\norm{f_i}}~, \\
        \hat{W}_j &= \frac{W_j}{\norm{W_j}}~, \\
        \hat{W}_j \cdot \hat{f}_i &= \cos(\hat{f}_i,~ \hat{W}_j)~.
    \end{aligned}
\end{equation}
After introducing a scaling term $s$, CosFace \cite{wang2018cosface} modifies the original softmax loss function into:
\begin{equation}
    \begin{aligned}
        L = -\log \frac{e^{s \cdot [\cos(\hat{f}_i,~ \hat{W}_{y_i})-m]}}{e^{s \cdot [\cos(\hat{f}_i,~ \hat{W}_{y_i})-m]} + \sum_{j \neq y_i} e^{s \cdot \cos(\hat{f}_i,~\hat{W}_j)}}~,
    \end{aligned}
    \label{eq:cosface}
\end{equation}
where $m$ is the margin for improving the decision boundary of face recognition networks, and $\hat{W}$ is the prototype of all identities in the training dataset.

\subsection{\MA}
\label{sec:mixfair}
To estimate the identity bias during training, we propose \MA~based on a mixing strategy. 

Given the two feature maps, $f_i$ and $f_j$, of two facial images from a biased face recognition network, the feature maps are assumed to be comprised of two terms, the bias-free representation and the identity bias term:
\begin{equation}
    f_i = r_i + b_i,~f_j = r_j + b_j~,
\end{equation}
where $r_i$ and $r_j$ are bias-free contour representations. $b_i$ and $b_j$ are their corresponding identity biases introduced by their races, genders, or other individual differences. 
Consider the mixed feature map of $f_i$ and $f_j$:
\begin{equation}
    f_m = \frac{1}{2} (f_i + f_j)~.
\end{equation}
When $f_i$ is a largely biased feature map, i.e., $\abs{b_i} \gg \abs{b_j}$, we observe that the output of a non-linear layer $M$ tend to preserve more similar feature of $f_i$:
\begin{equation}
    \label{eq:distance-compare}
    \cos(M(f_m), M(f_i))^2 - \cos(M(f_m), M(f_j))^2 = \epsilon > 0~,
\end{equation}
where $\cos$ indicates the cosine similarity function, and $\epsilon$ is the bias difference. We then infer which one of the feature maps has a larger identity bias according to $\epsilon$.

\paragraph{\MA.}
Instead of adopting adversarial training to remove the bias like previous works~\cite{debface, pass}, we propose a mixing strategy to remove the bias based on Equation~\eqref{eq:distance-compare}. Specifically, we firstly extract two intermediate feature maps of the facial images of two different identities from a network. The two feature maps are equally mixed together and the mixed feature map is then passed into the other layer, which we call ``debias layer'', to extract the final face feature map. Our \MA~is then established as the following:
\begin{equation}
    \begin{aligned}
        k_i = E(x_i),&~k_j = E(x_j)~,\\
        k_{mix} = \frac{1}{2}&(k_i + k_j)~,\\
        \epsilon_{i,j} = \cos(M(k_{mix}), M(k_i))&^2 - \cos(M(k_{mix}), M(k_j))^2~,
    \end{aligned}
    \label{eq:mixfair}
\end{equation}
where $(x_i, x_j)$ are two facial images, $E$ is an encoder, $M$ is the debias layer, and $\epsilon_{i,j}$ is the estimated bias difference between $x_i$ and $x_j$. When $k_i$ has large biases, $M(k_{mix})$ tends to preserve similar feature components of $M(k_i)$. Hence, the corresponding cosine similarity is larger than the one with $M(k_j)$, and vice versa. By making $\abs{\epsilon_{i,j}} \approx 0$, we can ensure that both $k_i$ and $k_j$ are not dominated by their own identity biases.

\subsection{MixFairFace Framework}
\label{sec:mixfairface}

To enforce networks to balance the biases of all training samples, i.e., $\abs{\epsilon_{i,j}} \approx 0$, we combine the CosFace loss function (Equation~\eqref{eq:cosface}) and the estimated bias difference from \MA~(Equation~\eqref{eq:mixfair}) as follows:
\begin{equation}
    \begin{aligned}
        L = -\log \frac{e^{s \cdot [\cos(\hat{f}_i,~ \hat{W}_{y_i})-m +\epsilon_{i,k}]}}{e^{s \cdot [\cos(\hat{f}_i,~ \hat{W}_{y_i})-m +\epsilon_{i,k}]} + \sum_{j \neq y_i} e^{s \cdot \cos(\hat{f}_i,~\hat{W}_j)}}~,
    \end{aligned}
    \label{eq:our-loss}
\end{equation}
where $i$ and $k$ denote two training samples from two different identities. We then minimize the identity bias difference by injecting $\epsilon_{i,k}$ into the prototype-based loss function. When $\epsilon_{i,k}$ is larger than zero, the identity bias of the training sample $i$ is larger than $k$. In this case, $(-m + \epsilon_{i,k})$ becomes larger to prevent our network from ``overly focusing'' on $i$, and vice versa. We have also tried to adopt $\abs{\epsilon_{i,k}}$ as an additional loss term, but our proposed loss function empirically achieves better performance.

\paragraph{MixFairFace.~}
The overall framework of MixFairFace is illustrated in Figure~\ref{fig:approach}. Following the previous work~\cite{cifp}, we adopt a ResNet-34~\cite{he2016deep} as the encoder to extract the intermediate feature maps of facial images, i.e., $E$ in Equation~\eqref{eq:mixfair}. Then, the debias layer consists of one fully-connected layer to extract the final feature vectors with 512 dimension, i.e., $M$ in Equation~\eqref{eq:mixfair}. Eventually, we have a final prototype layer that maps the 512 feature vectors to the predicted logits, and we optimize our network with the loss function in Equation~\eqref{eq:our-loss}. 
\section{Experiments}
\label{sec:experiments}
\begin{table*}[t]
  \centering
  \caption{The quantitative results trained on Balancedface~\cite{rl-rbn}. Note that the scale of TPR and FPR are 1e-2 and 1e-5, respectively.}
    \resizebox{\textwidth}{!}{\begin{tabular}{|r||cccccc|cccccc||c|}
    \hline
    \multirow{3}[6]{*}{\textbf{Method}} & \multicolumn{12}{c||}{\textbf{Attribute-based}}                                               & \multirow{2}[4]{*}{\textbf{Identity-based}} \bigstrut\\
\cline{2-13}          & \multicolumn{6}{c|}{\textbf{aTPR (1e-2)}}     & \multicolumn{6}{c||}{\textbf{aFPR (1e-5)}}    &  \bigstrut\\
\cline{14-14}          & \textbf{African} & \textbf{Asian} & \textbf{Caucasian} & \textbf{Indian} & \textbf{Avg} & \textbf{Std} & \textbf{African} & \textbf{Asian} & \textbf{Caucasian} & \textbf{Indian} & \textbf{Avg} & \textbf{Std} & \textbf{iFPR-std (1e-5)} \bigstrut\\
    \hline
    \hline
    \textbf{CosFace~\cite{wang2018cosface}} & 87.2  & 81.8  & 84.1  & 87.0  & 85.0  & 2.57  & 2.61  & 0.41  & 0.20  & 0.84  & 1.02  & 1.10  & 2.76 \bigstrut[t]\\
    \textbf{ArcFace~\cite{deng2019arcface}} & 85.5  & 79.5  & 81.3  & 85.0  & 82.8  & 2.90  & 2.45  & 0.48  & 0.19  & 0.90  & 1.01  & 1.01  & 2.61 \\
    \textbf{Mix (i)~\cite{mixup}} & 83.6  & 81.3  & 79.5  & 82.9  & 81.8  & 1.82  & 2.83  & 0.65  & 1.22  & 3.50  & 2.05  & 1.34  & 3.27 \\
    \textbf{Mix (m)~\cite{manifoldmixup}} & 87.4  & 81.7  & 83.5  & 86.6  & 84.8  & 2.66  & 2.68  & 0.43  & 0.21  & 0.71  & 1.01  & 1.13  & 2.73 \\
    \textbf{PASS~\cite{pass}} & 73.3  & 66.2  & 69.7  & 70.7  & 70.0  & 2.94  & 2.14  & 0.66  & 0.63  & 0.69  & 1.03  & \textbf{0.74} & 2.07 \\
    \textbf{CIFP~\cite{cifp}} & 88.4  & 83.3  & 85.2  & 88.5  & 86.4  & 2.55  & 2.54  & 0.44  & 0.20  & 0.86  & 1.01  & 1.06  & 2.71 \\
    \textbf{Ours} & 86.5  & 83.9  & 83.9  & 86.4  & 85.2  & \textbf{1.47} & 2.16  & 0.79  & 0.29  & 0.69  & 0.98  & 0.81  & \textbf{2.05} \bigstrut[b]\\
    \hline
    \end{tabular}}%
  \label{tab:balancedface}%
\end{table*}%
\begin{table*}[t]
  \centering
  \caption{The quantitative results trained on Globalface~\cite{rl-rbn}. Note that the scale of TPR and FPR are 1e-2 and 1e-5, respectively.}
    \resizebox{\textwidth}{!}{\begin{tabular}{|r||cccccc|cccccc||c|}
    \hline
    \multirow{3}[6]{*}{\textbf{Method}} & \multicolumn{12}{c||}{\textbf{Attribute-based}}                                               & \multirow{2}[4]{*}{\textbf{Identity-based}} \bigstrut\\
\cline{2-13}          & \multicolumn{6}{c|}{\textbf{aTPR (1e-2)}}     & \multicolumn{6}{c||}{\textbf{aFPR (1e-5)}}    &  \bigstrut\\
\cline{14-14}          & \textbf{African} & \textbf{Asian} & \textbf{Caucasian} & \textbf{Indian} & \textbf{Avg} & \textbf{Std} & \textbf{African} & \textbf{Asian} & \textbf{Caucasian} & \textbf{Indian} & \textbf{Avg} & \textbf{Std} & \textbf{iFPR-std (1e-5)} \bigstrut\\
    \hline
    \hline
    \textbf{CosFace~\cite{wang2018cosface}} & 87.5  & 87.8  & 84.5  & 88.9  & 87.2  & 1.88  & 2.17  & 1.15  & 0.06  & 0.65  & 1.01  & 0.89  & 2.64 \bigstrut[t]\\
    \textbf{ArcFace~\cite{deng2019arcface}} & 85.4  & 86.3  & 80.9  & 87.0  & 84.9  & 2.75  & 1.70  & 1.53  & 0.05  & 0.65  & 0.98  & \textbf{0.77} & 2.40 \\
    \textbf{Mix (i)~\cite{mixup}} & 83.1  & 84.2  & 80.8  & 84.9  & 83.3  & 1.79  & 2.69  & 0.96  & 0.05  & 0.38  & 1.02  & 1.18  & 3.33 \\
    \textbf{Mix (m)~\cite{manifoldmixup}} & 87.7  & 88.7  & 84.0  & 88.6  & 87.3  & 2.21  & 2.13  & 1.41  & 0.05  & 0.65  & 1.06  & 0.90  & 2.49 \\
    \textbf{PASS~\cite{pass}} & 77.9  & 73.6  & 74.8  & 77.7  & 76.0  & 2.14  & 2.87  & 0.56  & 0.16  & 0.47  & 1.02  & 1.25  & 2.76 \\
    \textbf{CIFP~\cite{cifp}} & 89.2  & 89.1  & 86.7  & 90.1  & 88.8  & 1.45  & 2.34  & 1.15  & 0.08  & 0.59  & 1.04  & 0.97  & 2.71 \\
    \textbf{Ours} & 87.4  & 87.9  & 86.2  & 89.3  & 87.7  & \textbf{1.28} & 1.95  & 1.37  & 0.08  & 0.65  & 1.01  & 0.82  & \textbf{2.09} \bigstrut[b]\\
    \hline
    \end{tabular}}%
  \label{tab:globalface}%
\end{table*}%

\begin{table*}[thb]
  \centering
  \caption{The quantitative results trained on MS1M~\cite{guo2016ms} and tested on IJB-C~\cite{IJB-C}. ``a1-a6'' indicate the attributes ``LightPink'', ``LightYellow'', ``MediumPink'', ``MediumYellow'', ``MediumDarkBrown'', and ``DarkBrown'', respectively. Note that the scale of TPR and FPR are 1e-2 and 1e-5, respectively.}
    \resizebox{\textwidth}{!}{\begin{tabular}{|r||cccccccc|cccccccc||c|}
    \hline
    \multirow{3}[6]{*}{\textbf{Method}} & \multicolumn{16}{c||}{\textbf{Attribute-based}}                                                                               & \multirow{2}[4]{*}{\textbf{Identity-based}} \bigstrut\\
\cline{2-17}          & \multicolumn{8}{c|}{\textbf{aTPR (1e-2)}}                     & \multicolumn{8}{c||}{\textbf{aFPR (1e-5)}}                    &  \bigstrut\\
\cline{18-18}          & \textbf{a1} & \textbf{a2} & \textbf{a3} & \textbf{a4} & \textbf{a5} & \textbf{a6} & \textbf{Avg} & \multicolumn{1}{c|}{\textbf{Std}} & \textbf{a1} & \textbf{a2} & \textbf{a3} & \textbf{a4} & \textbf{a5} & \textbf{a6} & \textbf{Avg} & \textbf{Std} & \textbf{iFPR-std (1e-5)} \bigstrut\\
    \hline
    \hline
    \textbf{CosFace~\cite{wang2018cosface}} & 73.4  & 71.6  & 79.5  & 70.3  & 64.7  & 71.8  & 71.9  & \textbf{4.79} & 0.18  & 1.00  & 0.52  & 1.09  & 1.35  & 2.93  & 1.18  & 0.96  & 3.68 \bigstrut[t]\\
    \textbf{CIFP~\cite{cifp}} & 73.1  & 72.3  & 80.1  & 69.9  & 64.7  & 71.4  & 71.9  & 5.00  & 0.28  & 0.88  & 0.34  & 1.21  & 1.47  & 2.90  & 1.18  & 0.96  & 3.64 \\
    \textbf{Ours} & 71.3  & 70.1  & 76.8  & 67.6  & 61.6  & 69.3  & 69.5  & 4.96  & 0.44  & 1.00  & 0.35  & 1.55  & 1.43  & 1.86  & 1.11  & \textbf{0.62} & \textbf{2.53} \bigstrut[b]\\
    \hline
    \end{tabular}}%
  \label{tab:ms1m}%
\end{table*}%

We conduct extensive experiments on two benchmark datasets, including Balancedface~\cite{rl-rbn} and Globalface~\cite{rl-rbn}. For fairness evaluation, we validate our framework and baselines on RFW~\cite{RFW} dataset. In addition, we provide the experimental results trained on a large-scale dataset, MS1M~\cite{guo2016ms}, and evaluated on IJB-C~\cite{IJB-C}.

\paragraph{Training and Evaluation Datasets.~} 
Balancedface provides around 1.3 million facial images with 7,000 identities for each race. Globalface provides around 2 million facial images with racial distribution following the real distribution on earth. RFW provides around 40,000 facial images with 3000 identities for each race. For experiments on large-scale datasets, we train our framework on MS1M, which provides facial images from 100K identities, and we adopt IJB-C as the testing dataset, which provides about 3,500 identities and the attributes labels of 6 skintones for evaluation. For IJB-C, we resampled it for making the number of identities in each attribute close.

\paragraph{Implementation Details.~}
We implement our approach with PyTorch~\cite{pytorch} framework. We apply Xavier~\cite{xavier} initialization and train our network with SGD optimizer which the momentum is set to 0.9 and weight decay is set to 5e-4. We train the network for totally 40 epochs with batch size 512. The learning rate starts from 0.1 and we apply learning rate decay with 0.1 decay factor on 8, 18, 30, and 34 epoch numbers. The scale $s$ and margin $m$ in Equation~\eqref{eq:our-loss} are set to 64 and 0.35. During training, we apply randomly horizontal flip to all training samples. All experiments in this paper are trained with 4x Tesla P100.

\subsection{Experimental Results}
In our experiments, we compare our MixFairFace with the following approaches. 1) ``CosFace'': the approach proposed in \cite{wang2018cosface}. 2) ``ArcFace'': the approach proposed in \cite{deng2019arcface}. 3) ``Mix (i)'': a baseline that adopts mixup~\cite{mixup} training and soft labels to train a face recognition network. 4) ``Mix (m)'': a baseline that adopts manifold-mixup~\cite{manifoldmixup} training and soft labels to train a face recognition network. 5) ``PASS'': the current state-of-the-art adversarial approach proposed in \cite{pass}. 6) ``CIFP'': the current state-of-the-art non-adversarial approach proposed in \cite{cifp} which improves the fairness of face recognition by minimizing the false positive rate inconsistency between different training samples.
\begin{figure}[thb]
    \centering
    \includegraphics[width=0.4\textwidth]{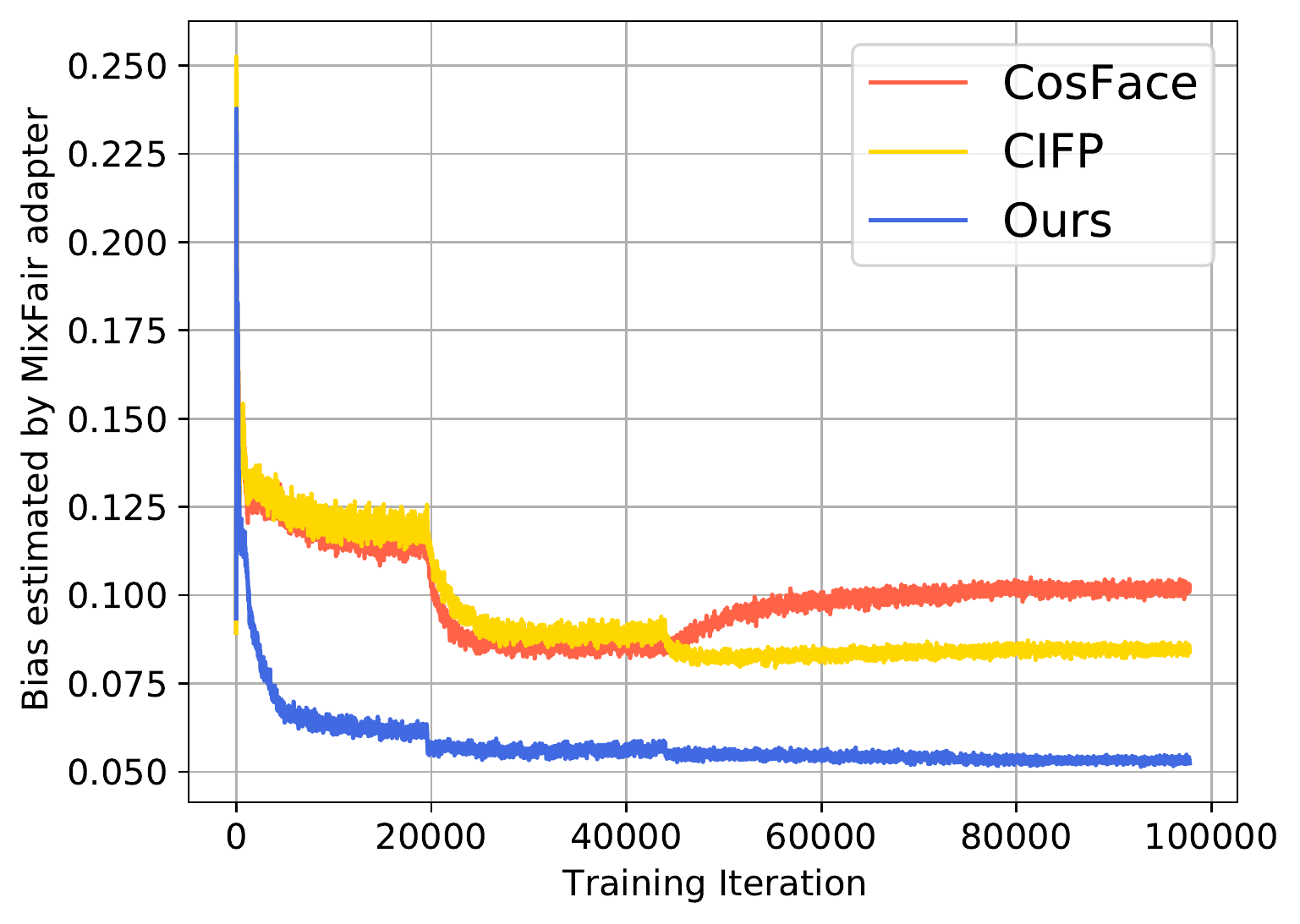}
    \caption{The change of bias difference estimated by Equation~\eqref{eq:mixfair} during training. We show the average absolute bias difference value of each training iteration. Our MixFairFace can significantly reduce the identity bias difference comparing to other approaches.}
    \label{fig:bias}
\end{figure}
\begin{figure*}[t]
    \centering
    \includegraphics[width=0.9\textwidth]{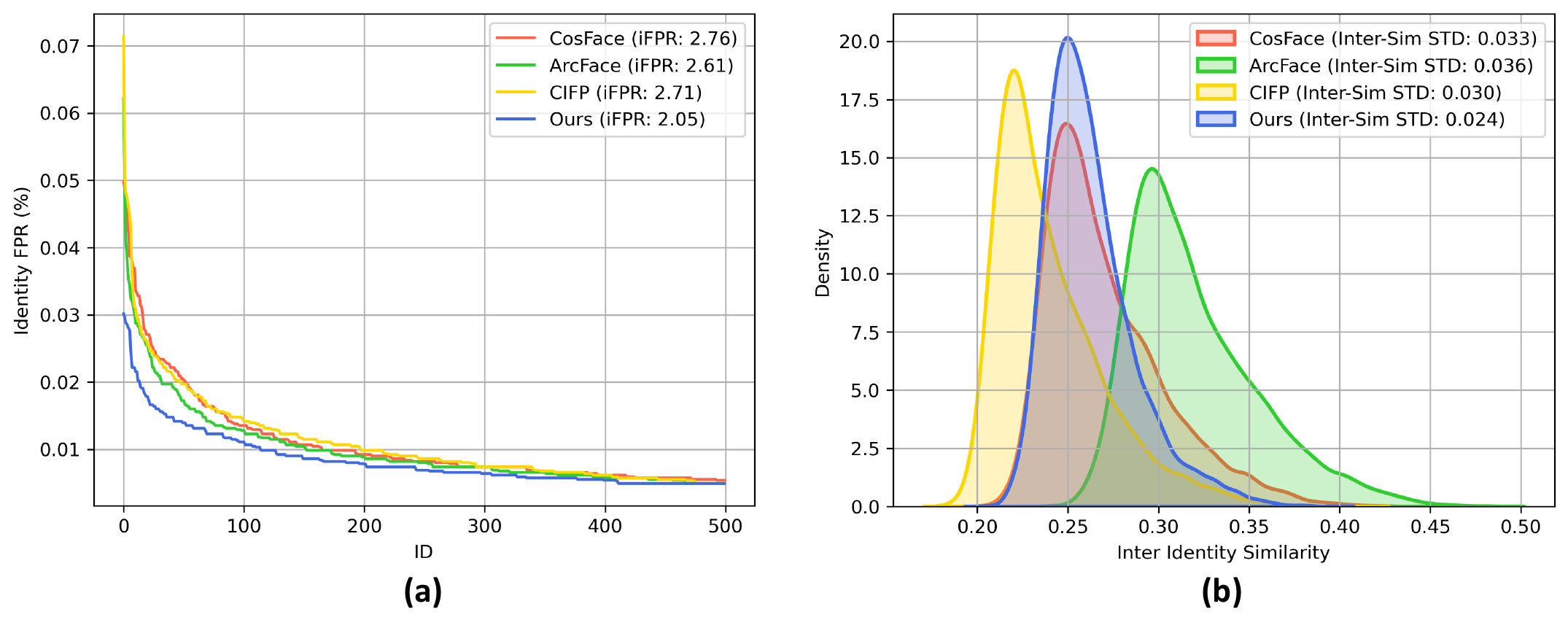}
    \caption{\textbf{(a)} The identity FPR (iFPR) distributions of different approaches trained on Balancedface. The iFPR is sorted from the highest to lowest on RFW identities, and we only show the top 500 identities with the largest FPR. Our MixFairFace significantly reduces the range between the lowest and highest FPR and thus we can achieve the smallest identity bias (iFPR-std). \textbf{(b)} The inter-identity similarity distributions of different approaches trained on Balancedface. Our MixFairFace achieves the largest density, i.e., the similarity variances across different identities are distributed more fairly than the other baselines. Thus, our approach can largely reduce identity bias differences.}
    \label{fig:compare-blancedface}
\end{figure*}

Our evaluation results of RFW trained on Balancedface and Globalface are shown in Table \ref{tab:balancedface} and \ref{tab:globalface}, respectively. We show the fairness performance by both attribute-based and identity-based metrics as described in Section~\ref{sec:eval}. 

\paragraph{General Comparison.} 
Compared with the baselines without any fairness constraints (``CosFace''), our MixFairFace achieves a slightly better average aTPR while improving the attribute-based standard deviation by 43\% on Balancedface. Compared with ``Mix (i)'', we found that directly mixing in the image spaces is harmful to the performance of face recognition tasks. As for mixing in the feature spaces (``Mix (m)''), we found directly applying manifold-mixup has no influence on both performance and fairness of face recognition.

\paragraph{Comparison with SOTA.}
Compared with the state-of-the-art adversarial approach (``PASS''), although the adversarial approach can improve the std of aFPR and the iFPR-std, the average aTPR decreases largely. Compared with the current state-of-the-art non-adversarial approach (``CIFP''), we found that CIFP is able to slightly improve the recognition performance and achieves the best average aTPR. However, the corresponding fairness improvement is highly limited and the iFPR-std is even worse than CosFace on Globalface. In general, our MixFairFace achieves the best aTPR-std and aFPR-std in attribute-based evaluation. For identity-based evaluation, our improvement of iFPR-std outperforms all other baselines, which demonstrates the proposed method significantly reduces the identity bias, i.e., the performance variance between different identities, by considering each identity fairly and balancing the bias difference between them by our \MA. 

\paragraph{Comparison on Large-Scale Datasets.}
We show the results trained on MS1M~\cite{guo2016ms} and tested on IJB-C~\cite{IJB-C} in Table~\ref{tab:ms1m}. Although the aTPR average of our approach is slightly worse than CosFace and CIFP, the aFPR std and iFPR-std of our approach still outperform the other approaches. For the most challenging attribute ``DarkBrown (a6)'', we can observe that the aFPRs of both CosFace and CIFP are extremely high, while our MixFairFace improves it by 36\% and the corresponding aTPR only decreases 3.5\%, which is an acceptable trade-off in practical scenarios. Hence, our MixFairFace framework shows great applicability for both medium and large scale datasets with an acceptable trade-off in the overall true positive rate.

\subsection{Discussion}
\paragraph{False Positive Rate Distribution.} 
To better illustrate the improvement of identity bias reduction by using our MixFairFace over other approaches trained on Balancedface, we show their FPR distributions in Figure~\ref{fig:compare-blancedface} (a). Our approach can largely reduce the difference between highest and lowest FPR identities, which shows that MixFairFace achieves the lowest identity bias difference between identities. \fuen{For the results trained on Globalface, please refer to our supplementary material.}

\paragraph{Inter-Identity Similarity Analysis.}
To investigate the fairness difference between the aforementioned approaches, we show their inter-identity cosine similarity distributions trained on Balancedface in Figure~\ref{fig:compare-blancedface} (b) based on Equation~\eqref{eq:similarity}. The distribution of our approach is more compact than those of other baselines, i.e., our density is the highest one. This indicates that the inter-similarity of each identity is closer and the identity bias is also much lower. Hence, our proposed framework can significantly improve the fairness in face recognition. \fuen{For the results trained on Globalface, please refer to our supplementary material.}

\paragraph{Identity Bias Difference.}
To verify the optimization of bias difference estimated by Equation~\eqref{eq:mixfair}, we show the average absolute bias difference of each training iteration in Figure~\ref{fig:bias}. Without any fairness constraints, we found the identity bias of ``CosFace'' starts to increase in the end. Compared to ``CIFP'', our MixFairFace significantly reduces the identity bias difference since our framework considers sample-to-sample fairness, and thus MixFairFace achieves the best fairness performance in face recognition.

\section{Conclusions}
In this paper, we propose MixFairFace framework to improve the fairness in face recognition. Instead of focusing on bias between demographic groups, we aim at solving the identity bias, i.e., the performance inconsistency between identities, such that we can avoid the skewness of demographic bias and achieve ultimate fairness results. To this end, we propose \MA~that adopts a mixing strategy to estimate the identity bias difference between two training samples. The estimated bias difference is then minimized by the proposed framework. In addition, we found that the evaluation protocol adopted by previous works highly underestimates the bias between demographic groups. Hence, we propose a new evaluation protocol based on both sensitive attributes and identities to provide a complete evaluation of fairness performance. We conduct experiments on several benchmark datasets, and our experiments demonstrate that our MixFairFace approach outperforms all other baselines and achieves state-of-the-art fairness performance. To compare the fairness performances of different approaches, we analyze the inter-identity similarity distribution and our framework achieves the highest density, which indicates our approach reduces the identity bias difference between identities and our identity bias training curve validates our formulation.

\paragraph{Acknowledgements.}
This work is a research project in Microsoft, and also supported by Ministry of Science and Technology of Taiwan (MOST 110-2634-F-002-051 and Taiwan Computing Cloud).


\end{document}